\documentclass[10pt,twocolumn,letterpaper]{article}

\usepackage[pagenumbers]{cvpr} 

\usepackage{multirow}
\usepackage[table]{xcolor}

\usepackage{enumitem}
\usepackage{amssymb}
\usepackage{booktabs}
\usepackage{amsmath}
\usepackage{longtable}

\definecolor{cvprblue}{rgb}{0.21,0.49,0.74}
\usepackage[pagebackref,breaklinks,colorlinks,allcolors=cvprblue]{hyperref}

\def\paperID{15944} 
\def\confName{CVPR}
\def\confYear{2026}

\title{PA-MIL: Phenotype-Aware Multiple Instance Learning Guided by Language Prompting and Genotype-to-Phenotype Relationships}

\author{Zekang Yang$^{1}$ \quad Hong Liu$^{1}$\thanks{Corresponding author: hliu@ict.ac.cn} \quad Xiangdong Wang$^{1}$  \\
$^{1}$ Institute of Computing Technology, Chinese Academy of Sciences}

\begin{document}
\maketitle
\begin{abstract}
Deep learning has been extensively researched in the analysis of pathology whole-slide images (WSIs).
However, most existing methods are limited to providing prediction interpretability by locating the model’s salient areas in a post-hoc manner, failing to offer more reliable and accountable explanations.
In this work, we propose Phenotype-Aware Multiple Instance Learning (PA-MIL), a novel ante-hoc interpretable framework that identifies cancer-related phenotypes from WSIs and utilizes them for cancer subtyping.
To facilitate PA-MIL in learning phenotype-aware features, we 1) construct a phenotype knowledge base containing cancer-related phenotypes and their associated genotypes. 2) utilize the morphological descriptions of phenotypes as language prompting to aggregate phenotype-related features. 3) devise the Genotype-to-Phenotype Neural Network (GP-NN) grounded in genotype-to-phenotype relationships, which provides multi-level guidance for PA-MIL.
Experimental results on multiple datasets demonstrate that PA-MIL achieves competitive performance compared to existing MIL methods while offering improved interpretability.
PA-MIL leverages phenotype saliency as evidence and, using a linear classifier, achieves competitive results compared to state-of-the-art methods.
Additionally, we thoroughly analyze the genotype-phenotype relationships, as well as cohort-level and case-level interpretability, demonstrating the reliability and accountability of PA-MIL.
The code is avaliable at \url{https://github.com/yang-ze-kang/PA-MIL}

\end{abstract}    
\section{Introduction}
Deep learning technology has demonstrated outstanding performance in the automatic analysis of pathology whole slide images (WSIs), significantly advancing Computational Pathology (CPath) and cancer-assisted diagnosis~\cite{clam, chen2022pan, transmil, chen2024wsicaption, xu2024multimodal, li2024generalizable}.
Although deep neural network models can achieve excellent performance, their black-box nature makes it difficult to provide reasonable interpretability for their predictions. However, in real clinical contexts, a reliable and accountable AI algorithm is extremely important. It should provide more detailed evidence to support its predictions in order to gain the trust of users (or pathologists).
\begin{figure}[t]
    \centering
    \includegraphics[width=0.47\textwidth]{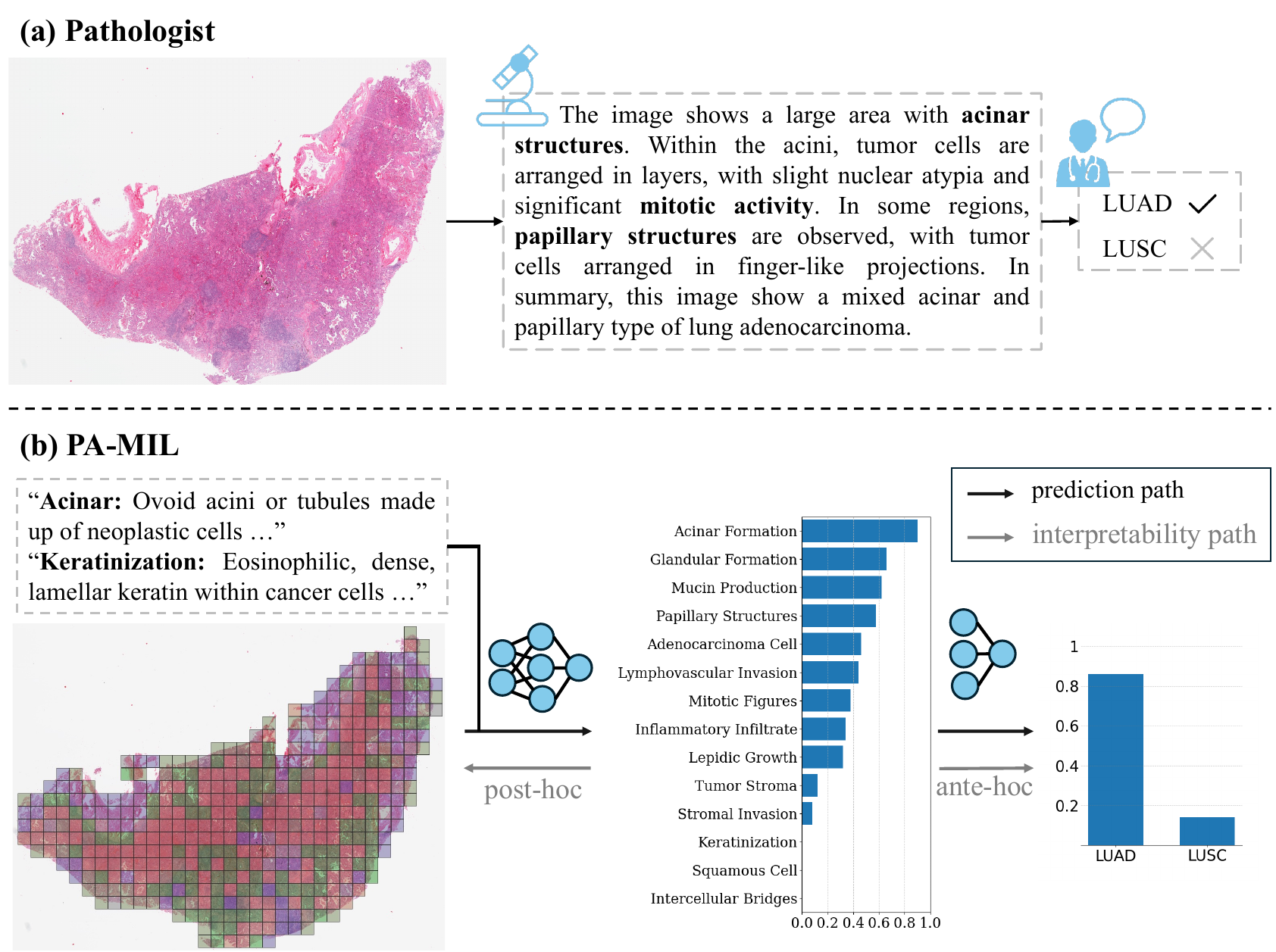}
    \caption{Comparison of the diagnostic process. (a) The diagnostic process of  pathologists. (b) The predictive pathway and diagnostic evidence of PA-MIL. }
    \label{fig:teaser}
    \vspace{-3mm}
\end{figure}
Previous post-hoc interpretability methods~\cite{clam, chen2022scaling, song2024morphological, jaume2024modeling} are limited to locating salient regions without providing more detailed or convincing diagnostic evidence.
In practice, as shown in Figure~\ref{fig:teaser}a, pathologists analyze WSIs by observing cell morphology and tissue structures to identify the clinical saliency of specific phenotypes (e.g., glandular structures, keratinization, intercellular bridges). They then make diagnoses based on the correlation between these phenotypes and cancer subtypes.
It naturally raises the question:
\textit{Can a deep neural network model act like pathologists, identifying cancer phenotypes from WSIs and making diagnoses based on them?}

There are two main challenges that hinder the emergence of such a model.
(1) \textit{Which phenotypes to recognize from pathological images?}
(2) \textit{How can neural network models learn phenotype-related features from WSIs without phenotype annotations?}
To address these challenges, we first build a phenotypic knowledge base by leveraging GPT-4 and the expertise of pathologists.
Subsequently, we introduce textual descriptions of phenotypes and transcriptomics data paired with WSIs to assist in learning cancer phenotype features from WSIs.
Due to the superpixel nature of pathology images, they are often divided into patches for processing. In recent years, pathology-specific CLIP models~\cite{huang2023visual, zhang2023biomedclip, lu2024visual, sun2024pathgen} have demonstrated exceptional performance in tasks such as text-to-image retrieval. Such CLIP models can be used to roughly identify patches related to specific phenotypes from the large number of patches in each WSI, based on language prompting of the phenotypes.
Compared to a single subtype label, the transcriptomic data offer a much richer and more detailed molecular profile.
Furthermore, there are close relationships between genotypes and phenotypes, as gene expression at the microscopic level determines the macroscopically visible phenotypes~\cite{costanzo2019global, kim2016understanding}.
Based on the genotype-to-phenotype relationships, finer-grained genotype-specific information can be provided to the neural network models to help them recognize phenotypes from histological images.

Based on these insights, we propose a phenotype-aware multiple instance learning model (PA-MIL) that can identify phenotypes from WSIs and make diagnoses based on them. 
PA-MIL first utilizes a text encoder and an image encoder to extract textual features of phenotype descriptions and image features of WSIs, respectively. It then utilizes the cross-attention between the textual features and patch features to obtain phenotype features. Next, the clinical saliency of each phenotype is predicted based on its features, and finally, cancer diagnosis is performed based on the clinical saliency of the phenotypes.
To facilitate model training, we also devise a genotype-to-phenotype neural network model (GP-NN) based on transcriptomic data and the relationship between genotypes and phenotypes as the teacher model.
GP-NN shares a similar design with PA-MIL. It first divides the transcriptomic data into different groups based on their associations with various phenotypes. Then, different MLPs are used to learn the interactions between genes within each group, generating corresponding genotype-to-phenotype features. These features are then used to predict the clinical saliency of the associated phenotypes. Finally, the cancer diagnosis is made based on the clinical saliency of each phenotype.
Due to the similar design of GP-NN and PA-MIL, GP-NN can provide PA-MIL with multi-hierarchical supervisory information, including phenotype feature level and phenotype saliency level.
This helps PA-MIL learn more precise phenotype-related information during the training phase.
The contributions of this work are summarized as follows:

\begin{itemize}
    \item  We construct a phenotype knowledge base that includes cancer-related phenotypes observable in WSIs. Each phenotype contains its morphological description and the associated gene set.
    
    \item  We propose a phenotype-aware multiple instance learning method (PA-MIL), the first method that explicitly identifies the cancer-related phenotypes from WSIs and uses phenotypic evidence for diagnosis prediction.
    
    \item  We propose a phenotype learning approach that leverages language prompts and genotype-to-phenotype relationships to guide PA-MIL in extracting discriminative phenotypic features from WSIs.
    
    \item  We conduct comprehensive experiments and interpretability analyses, demonstrating that PA-MIL can achieve competitive performance compared to previous MIL methods and provide more clinically valuable phenotype-based evidence for its predictions.

\end{itemize}

\section{Related Work}
\subsection{Multiple Instance Learning}
Early methods for pathological image analysis primarily relied on pathologists manually annotating regions of interest~\cite{mobadersany2018predicting, zhu2016deep, wang2014novel, yao2016imaging}.
Subsequent research introduced multiple instance learning techniques, enabling the direct analysis of entire pathological images~\cite{amil, dsmil, transmil, yang2024scmil, chen2022pan}.
Existing MIL methods can be categorized into two groups based on the following principles: the instances within a bag are independent, or the instances within a bag are interdependent.
The former primarily focuses on identifying key instances from a large number of patches in a whole slide image (WSI)~\cite{campanella2019clinical, amil, shi2020loss, qu2022dgmil}.
The latter mainly emphasizes learning the relationships between instances using methods such as prototype learning~\cite{yao2020whole, yu2023prototypical, vu2023handcrafted, song2024morphological}, graph neural networks~\cite{chen2021whole, guan2022node, hou2023multi, chan2023histopathology}, and self-attention mechanisms~\cite{transmil, dsmil, fourkioti2023camil, li2024rethinking}.
PA-MIL is somewhat similar to prototype-based MIL methods, as both leverage the distribution of different phenotypes or prototypes within a specific WSI as the basis for diagnosis. 
However, prototypes are typically obtained through global clustering and lack explicit semantic meaning.
While post-hoc analysis can reveal that some prototypes contain specific phenotype patterns, they still lack stronger reliability and interpretability.
In contrast, cancer-related phenotypes are directly extracted from WSIs based on histopathological knowledge, providing clear semantic meaning and stronger interpretability.
\begin{figure*}[!thbp]
    \centering
    \includegraphics[width=0.96\textwidth]{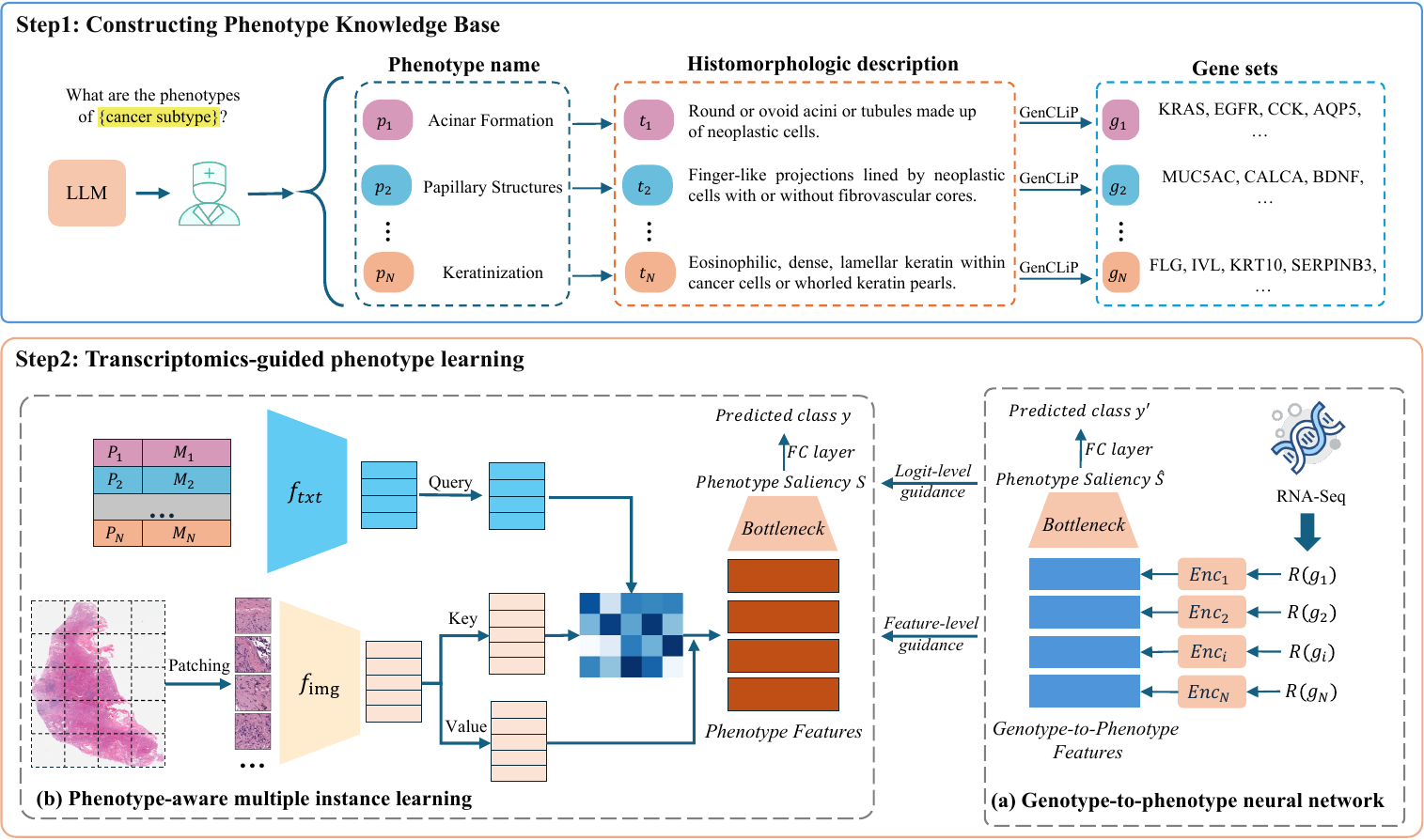}
    \caption{Overview of proposed PA-MIL and training method. \textbf{Step 1:} Constructing phenotypic knowledge base. \textbf{Step 2a:} Training genotype-to-phenotype neural network. \textbf{Step 2b:} Training phenotype-aware multiple instance learning with transcriptomic guidance.}
    \label{fig:overview}
\end{figure*}
\subsection{Interpretability Methods}
Methods based on hand-crafted features~\cite{hamilton1994expert, keenan2000automated, zamanitajeddin2021cells} first extract properties related to the morphology and spatial distribution of cells in WSIs. These features are then used for diagnostic predictions, offering good interpretability.
Deep learning-based methods demonstrate superior performance but lack interpretability, relying on contributions of attention~\cite{amil, transmil} or gradient~\cite{selvaraju2017grad, zheng2022graph} to identify salient regions of WSI.
Such post-hoc methods merely highlight superficial associations between inputs and outputs, without offering deeper or more convincing evidence for predictions. SI-MIL~\cite{SI-MIL} compromises between model performance and interpretability by identifying salient regions through post-hoc methods and using hand-crafted features from these regions for diagnostic predictions.
Our goal is to design a self-interpretability deep neural network model that is capable of identifying cancer-related phenotypes and making diagnoses based on these phenotypes.
In the domain of natural images, the Concept Bottleneck Models (CBMs)~\cite{koh2020concept} identify human-understandable concepts from the input and make predictions based on these interpretable concepts, providing strong interpretability for the final prediction.
However, CBMs are limited by their reliance on concept-annotated data and a compromise in performance.
In this paper, we introduce language prompting and genomic data to address these limitations.

\section{Method}
In this section, we present the architecture and training methods of Phenotype-Aware Multiple Instance Learning (PA-MIL) in detail (overview \Cref{fig:overview}).
We first construct a phenotypic knowledge base containing cancer-related phenotypes and their associated genotypes (Sec.~\ref{sec:phenotype_knowledge_base}).
Next, we provide a detailed explanation of the proposed PA-MIL (Sec.~\ref{sec:pamil}).
Based on the relationships between phenotypes and genotypes, we devise a genotype-to-phenotype neural network (GP-NN) using transcriptomic data (Sec.~\ref{sec:gpnn}).
Finally, we introduce a method to use transcriptome information to guide PA-MIL to learn cancer-related phenotypes from WSIs. (Sec.~\ref{sec:genelamp}).

\subsection{Phenotype Knowledge Base}
\label{sec:phenotype_knowledge_base}
First, we construct a phenotype knowledge base $\mathcal{K}$ to provide medical prior knowledge to support our method.
Each phenotype in $\mathcal{K}$ has its corresponding histomorphological description and function-related gene set.
In this section, we provide a detailed description of the process of constructing the phenotype knowledge base.
\textbf{Identifying cancer-related phenotypes:}
Large language models (LLMs) possess a vast amount of knowledge. We prompt LLMs to generate potential phenotypes associated with specific cancers, which are then refined and validated through expert correction.
As a result, we identify $N$ phenotypes $\mathcal{P}=\{p_1,p_2,...,p_{N}\}$ related to the specific cancer $c$.
\textbf{Generating histomorphological-related descriptions:}
In a similar manner, we design prompts for LLMs to describe the morphological structures of each phenotype and manually correct their outputs, resulting in the histomorphological descriptions $\mathcal{T}=\{t_1,t_2,...,t_{N}\}$ for all phenotypes.
\textbf{Retrieving function-related gene sets:}
GenCLiP~\cite{wang2020genclip} is a tool that provides rapid retrieval of function-related genes according to custom terms from the entire PubMed database.
We use the names and synonyms of the phenotypes as queries for GenCLiP and obtain the gene sets $\mathcal{G}=\{g_1,g_2,...,g_{N}\}$ associated with the phenotypes.
Then we have the phenotype knowledge base $\mathcal{K}=\{(p_i,t_i,g_i)|i\in I, p_i\in\mathcal{P},t_i\in\mathcal{T},g_i\in\mathcal{G}\}$.

\subsection{Phenotype-Aware Multiple Instance Learning}
\label{sec:pamil}
\textbf{Multiple Instance Learning (MIL).}
MIL is a weakly-supervised learning framework designed for handling data with a set-based structure.
When applying MIL to computational pathology, a given WSI $X^j$ is segmented into multiple fixed-size patches $X^j=\{x_i^j\}_{i=1}^{M_j}$, and a pre-trained feature extractor $f_{img}$ is used to extract features $H^j=f_{img}(X^j)$ from these patches.
Where $M_j$ denotes the number of patches extracted from $X^j$.
The set of features $H^j\in\mathbb{R}^{M_j\times d}$ is aggregated to obtain a WSI-level feature $h_{w}^j\in\mathbb{R}^{d}$. And $h_{w}^j$ is then applied to prediction tasks.

The PA-MIL we propose aims to learn a WSI-level feature $S^j=\{s_i^j\}_{i=1}^{N}$ with $S^j\in\mathbb{R}^{N}$ that possesses inherent interpretability, where each element $s_i^j$ corresponds to the saliency score of a cancer-related phenotype.
And the diagnostic predictions are made based on $S^j$.
In each inference, PA-MIL takes the names $\mathcal{P}$ and histomorphological descriptions $\mathcal{T}$ of phenotypes and the features $H^j$ of WSI $X^j$ as input.
\textbf{Encoding language prompting:} We use a weight-frozen text feature extractor $f_{txt}$ to extract the textual features $U=\{u_i\}_{i=1}^{N}$ with $u_i\in\mathbf{R}^d$ for all phenotypes:
\begin{equation}
    u_i=f_{txt}([p_i,t_i]); \quad i\in\{1,2,...,N\}
\end{equation}
\textbf{Extracting phenotype-related features:} Considering that pretrained vision-language models have limited ability to recognize cancer-related phenotypes~\cite{sun2025label}, we design a learnable cross-attention method to fine-tune it and extract phenotype-related features $V^j\in\mathbb{R}^{N\times d}$ from WSI:
\begin{equation}
    \begin{split}
        V^j&=f_{p}(H^j,U) \\
        &=A^j\times W_v^T(H^j) \\
        &=\text{softmax}(\frac{W_q(U)W_k^T(H^j)}{\sqrt{d}})W_v^T(H^j)
    \end{split}
\end{equation}
Where $W_q(\cdot)$,$W_k(\cdot)$ and $W_v(\cdot)$ are learnable linear projections.
$A^j\in\mathbb{R}^{N\times M_j}$ represents the contribution of each patch in $X^j$ to different phenotypes.
\textbf{Aligning cohort-level phenotype features:}
PA-MIL aims to extract mutually independent phenotypic features.
Accordingly, features representing the same phenotype across different samples within a cohort should be close in the feature space, while features of different phenotypes should be well separated.
Assuming the cluster centers of phenotypes are represented by $\overline{V}\in\mathbb{R}^{N_c\times d}$. For the phenotype feature $V^j_i$ , it has the positive sample $\overline{V}_i$ and negative samples $\{\overline{V}_k\}_{k\neq i}^{N}$. Then we have the objective function:
\begin{equation}
    L_{contrast} = \frac{1}{N}\sum_{k=1}^{N}log(\frac{exp(V^j_k\overline{V}_k^T/\tau)}{\sum_{k'}^{N}exp(V^j_{k}\overline{V}_{k'}^T/\tau)})
\end{equation}
Due to the superpixels of WSIs, it is difficult to compute $\overline{V}$ by using a large batch size. We draw on the previous contrastive learning method based on self-distillation and use the momentum update strategy to obtain the cluster centers:
\begin{equation}
 \overline{V} = \alpha\overline{V}+(1-\alpha)V^j
\end{equation}
where $\alpha$ denotes the momentum value.
\textbf{Predicting saliency scores of phenotypes:} To facilitate better interpretability, we introduce a bottleneck~\cite{koh2020concept} layer $W_w(\cdot)$ to predict the saliency scores of the phenotypes.
To mitigate the leakage of subtype label information into phenotypic prediction, we employ a layer normalization (LN) layer as the activation function to compute relative saliency scores, which are then used as the final saliency scores
$S^j=\text{LN}(W_w(V^j))$.
\textbf{Classifying cancer subtype:} Histomorphological phenotypes are important evidence for cancer diagnosis and subtype classification. We use the saliency scores of phenotypes as the input features for the classifier $f_a(\cdot)$ to predict cancer subtypes $y_w^j$ and use the cross-entropy loss function $L_{ce}=y_t^jlog(y_w^j)$ as the objective function for end-to-end training, where $y_t^j$ is the label of the j-th sample.

\subsection{Genotype-to-Phenotype Neural Network}
\label{sec:gpnn}
Phenotypes are potentially associated with specific gene sets, and the phenotypes observed in histopathological images are typically the macroscopic results of specific gene expressions at the microscopic level~\cite{zhang2023molecular}.
Transcript abundance (RNA-Seq) reflects gene expression levels. Therefore, we devise the genotype-to-phenotype neural network (GP-NN) to learn phenotype-related information from transcriptomics data.
\textbf{Encoding phenotype-related features:}
We first divide all the RNA-Seq abundance $R^j$ into different sets $\{R^j(g_1),R^j(g_2),...,R^j(g_{N})\}$ based on the relationship between genotype and phenotype, where $j$ denotes the j-th sample.
RNA-Seq abundance is typically quantified as an $1\times1$ attribute.
We can employ a set of multi-layer perceptions (MLPs) $\{Enc_1,Enc_2,...,Enc_{N}\}$ to encode the interactions between genes within different gene sets that contribute to the corresponding phenotypes, thereby obtaining phenotype-related features $Z^j\in\mathbb{R}^{N\times d}$:
\begin{equation}
    z_i^j = Enc_i(R^j(g_i)); \quad i\in\{1,2,...,N\}
\end{equation}
\textbf{Predicting the saliency scores of phenotypes:}
Similar to PA-MIL, we also employ a bottleneck layer $W_g(\cdot)$ followed by a LN layer to predict the saliency scores $\hat{S}^j\in\mathbb{R}^{N}$ of the phenotypes:
$\hat{S}^j=\text{LN}(W_g(Z^j))$.
\textbf{Cancer subtyping:} Finally, the scores $\hat{S}^j$ are fed into the classifier $f_b(\cdot)$ to predict cancer subtypes $\hat{y_g}^j=f_b(\hat{S}^j)$, thereby supervising the training of the GP-NN.

\subsection{Transcriptomics-Guided Phenotype Learning}
\label{sec:genelamp}
To facilitate PA-MIL in learning phenotype-related information, we propose a transcriptomics-guided phenotype learning method.
In the following content, we provide a detailed introduction to the design of the guidance methods and learning methods.

\textbf{Guidance method.}
We utilize the weight-frozen GP-NN to provide both logit-level and feature-level guidance for PA-MIL, enhancing its interpretability while achieving competitive performance on downstream tasks.
\textbf{Logit-level:} Minimize the distance $L_{logit}$ between the phenotype saliency scores predicted by PA-MIL and GP-NN.
\textbf{Feature-level:} Minimize the distance $L_{feat}$ between the phenotype-related features extracted by PA-MIL and GP-NN.
We use L2 distance as the distance metric function:
\begin{equation}
    L_{logit} = \frac{1}{N}\|S^j-\hat{S}^j\|_2^2; \quad
\end{equation}
\begin{equation}
    L_{feat} = \frac{1}{N}\sum_{i=1}^{N}\frac{1}{d}\|v_i^j-z_i^j\|_2^2
\end{equation}

\begin{table*}[thbp]
    \centering
    \caption{Results of PA-MIL and baselines on four different datasets. All methods except $\ast$ use CONCH~\cite{lu2024visual} to extract path features. $\Diamond$ denotes post-hoc methods and $\heartsuit$ denotes ante-hoc methods. The mean and standard deviation of the five-fold cross-validation are reported. Best performance in \textbf{bold}, second best \underline{underlined}.}
    \scalebox{0.86}{
    \begin{tabular}{l|cc|cc|cc|cc}
    \toprule
    Train on                 & \multicolumn{4}{c|}{TCGA-NSCLC}                             & \multicolumn{4}{c}{TCGA-RCC}      \\ \cline{2-9} 
    \multirow{2}{*}{Test on} & \multicolumn{2}{c|}{TCGA} & \multicolumn{2}{c|}{CPTAC} & \multicolumn{2}{c|}{TCGA} & \multicolumn{2}{c}{CPTAC} \\
                             & Acc.         & AUC          & Acc.          & AUC         & Bal. acc.         & F1         & Bal. acc.         & F1         \\
    \midrule
    $\Diamond$ AMIL\cite{amil}          & 94.19±1.64 & 98.01±0.98 & 91.98±0.56 & 97.60±0.27 & 94.24±4.25 & \underline{95.77±1.87} & 80.57±1.59 & 92.62±0.56 \\
    $\Diamond$ TransMIL~\cite{transmil} & 93.60±1.28 & 97.09±2.71 & 91.17±0.74 & 96.60±0.65 & 93.62±2.64 & 95.19±0.87 & 83.92±5.32 & 91.20±1.26 \\
    $\Diamond$ DSMIL~\cite{dsmil}       & 93.62±1.34 & 97.94±1.19 & 91.60±0.43 & 97.46±0.05 & 93.92±1.93 & 94.68±0.56 & 82.19±0.77 & 94.01±0.37 \\
    $\Diamond$ AttMISL~\cite{AttMISL}       & 92.93±1.11 & 97.82±0.93 & 91.34±0.24 & 97.48±0.13 & 94.44±1.70 & 95.58±1.38 & 80.51±2.27 & 91.77±0.82 \\
    $\heartsuit$ AdditiveMIL~\cite{AdditiveMIL}& 94.09±1.72 & 98.02±1.08 & 91.92±0.64 & 97.41±0.29 & 95.42±1.51 & 95.68±0.70 & 81.86±1.21 & 92.76±0.66 \\
    $\heartsuit$ SI-MIL$^{\ast}$\cite{SI-MIL} & 88.4 & 94.1 & - & - & - & - & - & - \\
    $\heartsuit$ Panther+lin.~\cite{song2024morphological}       & 92.27±2.68 & 96.92±1.36 & 89.90±0.66 & 96.87±0.30 & 90.34±5.31 & 94.41±0.91 & 75.60±1.35 & 92.27±0.63 \\
    $\heartsuit$ Panther+MLP~\cite{song2024morphological}        & 92.85±2.15 & 97.25±1.37 & 90.51±1.13 & 97.20±0.29 & 91.05±3.94 & 93.92±1.21 & 78.88±3.28 & 92.33±0.80 \\
    \midrule
    $\heartsuit$ PA-MIL$_S$(feature) & \textbf{94.77±1.56} & 97.80±0.91 & 92.39±0.18 & \underline{97.77±0.15} & 94.52±1.78 & 95.26±0.71 & \underline{84.36±3.14} & 92.97±1.58 \\
    $\heartsuit$ PA-MIL$_J$(feature) & 94.09±1.71 & \underline{98.17±0.99} & \textbf{92.96±0.48} & \textbf{97.95±0.11} & \textbf{96.00±0.88} & \textbf{95.63±0.85} & 82.22±1.31 & \underline{94.06±0.48} \\
    \rowcolor{Orange!20}
    $\heartsuit$ PA-MIL$_S$(score) & 94.19±1.71 & 98.07±0.84 & \underline{92.71±0.46} & 97.75±0.20 & 93.03±3.84 & 94.52±0.75 & \textbf{84.90±1.93} & \textbf{94.50±0.78} \\
    \rowcolor{Orange!20}
    $\heartsuit$ PA-MIL$_J$(score) & \underline{94.68±1.30} & \textbf{98.42±0.84} & 92.62±0.80 & 97.69±0.50 & \underline{95.66±1.86} & 95.69±0.59 & 82.58±0.85 & 93.34±1.01 \\
    \bottomrule
    \end{tabular}}\label{tab:sota}
    \vspace{-3mm}
\end{table*}

\textbf{Learning method.} We systematically study two training methods.
\textbf{Sequential training:} PA-MIL first learns to extract phenotype-related features and predict phenotype saliency scores:
$\hat{f_{p}},\hat{W_w} = \text{arg min}_{f_{p},W_w}\sum_j(L_{feat}+L_{logit}+L_{contrast})$.
Subsequently, with the weights of $f_p$ and $W_w$ frozen, PA-MIL learns to predict cancer subtypes based on the phenotype saliency scores:
$\hat{f_{a}} = \text{arg min}_{f_a}\sum_jL_{ce}$.
\textbf{Joint training:} Treat phenotype learning as an auxiliary branch to enable simultaneous learning of subtype classification and phenotype prediction: $\hat{f_{a}}, \hat{f_{p}},\hat{W_w} = \text{arg min}_{f_a,f_{p},W_w}\sum_j[L_{ce}+L_{contrast}+\lambda(L_{feat}+L_{logit})]$
, where $\lambda$ denotes the weight for phenotype learning supervision.

\section{Experiment and Reseults}
\subsection{Dataset and Implementation Details}
\textbf{Datasets.} We evaluate PA-MIL on two different subtyping tasks and four different WSI datasets: Non-Small Cell Lung Carcinoma (NSCLC) subtyping on
TCGA (2 classes) and CPTAC (2 classes). Renal Cell Carcinoma (RCC) subtyping on TCGA (3 classes) and CPTAC (2 classes).
We train the models using datasets from the TCGA and then test them on datasets from both the TCGA and CPTAC.
For TCGA, we perform 5-fold cross-validation to conduct five repeated experiments.
The datasets from CPTAC were used as an external dataset to validate the model’s generalizability.
We use the accuracy and area under the curve (AUC) metrics for evaluation on NSCLC subtyping and balanced accuracy and weighted F1 as evaluation metrics for RCC subtyping due to the imbalance among subtypes.

\textbf{Implementation.}
We follow the CLAM~\cite{clam} to preprocess the WSIs.
We first employ OTSU's threshold method to segment the tissue regions in the WSIs at low magnification.
Then we divide the WSIs at 20x magnification into non-overlapping patches of size 512$\times$512 pixels, which are subsequently resized to 448$\times$448 pixels before feature extraction.
We set the momentum value $\alpha$ at 0.9.
All the comparison models except SI-MIL~\cite{SI-MIL} utilize the image encoder of CONCH~\cite{lu2024visual} to extract the patch features.
We also use the text encoder of CONCH to extract the textual features of the phenotypes.
All models are trained for 20 epochs using the Adam optimizer~\cite{kingma2014adam}, with a batch size of 1 and a gradient accumulation step of 32. The learning rate is set to $5\times10^{-4}$, the weight decay to $1\times10^{-4}$, and the gradient clipping threshold to 10.

\subsection{Comparisons with State-of-the-Art}

\textbf{Baselines.}
We compare our method with both post-hoc methods and ante-hoc methods.
AMIL~\cite{amil}, TransMIL~\cite{transmil}, DSMIL~\cite{dsmil}, and AttMISL~\cite{AttMISL} lack intrinsic interpretability and can only be explained through post-hoc methods.
AdditiveMIL~\cite{AdditiveMIL}, SI-MIL~\cite{SI-MIL}, and Panther~\cite{song2024morphological} optimize the network design in an effort to provide ante-hoc explanations for the model’s predictions.
AMIL is used as the base model for AdditiveMIL.
SI-MIL does not provide the complete hand-crafted features, and we directly report the results from their paper.
Panther+lin. and Panther+MLP predict based on prototype-related features, using a linear classifier and a multilayer perceptron, respectively.

\textbf{Results.} \Cref{tab:sota} presents the result. PA-MIL$_{S}$ and PA-MIL$_{J}$ represent the models obtained through sequential learning and joint learning, respectively.
“Feature” refers to using a transformer for prediction based on phenotypic features.
“Score” represents using a linear classifier for prediction based on phenotypic saliency scores.
$\ast$ are obtained from their papers, and other results are re-implemented with the same image encoder.
Overall, both PA-MIL$_{S}$ and PA-MIL$_{J}$ consistently outperform
or are on par with all post-hoc and ante-hoc baselines.

\textbf{Phenotypic features \textit{vs.} phenotypic saliency scores} Due to the bottleneck layer, a saliency score loses more information compared to a feature vector. Additionally, the linear classifier can learn fewer patterns compared to the nonlinear classifier.
However, subtyping with phenotypic saliency scores has achieved performance that is comparable to, or even better than, phenotypic features.
It suggests that the phenotypic saliency scores predicted by PA-MIL is directly associated with cancer subtypes, and excellent results can be achieved with a simple linear classifier.
In contrast, Panther+MLP outperforms Panther+lin, suggesting that prototypes obtained through clustering may not be directly related to the subtype classification.
\begin{table}[thbp]
    \centering
    \caption{Ablation of modules. `LP' means language prompting. `PA' means cohort-level phenotypic features alignment. `TG' means transcriptomics guidance.}
    \begin{tabular}{lcccc}
    \toprule
    \multirow{2}{*}{Module} & \multicolumn{2}{c}{TCGA} & \multicolumn{2}{c}{CPTAC} \\ \cmidrule(lr){2-3} \cmidrule(lr){4-5} 
                          & Acc         & AUC        & Acc         & AUC         \\ \midrule
    baseline              & 94.09       & 97.13      & 91.80       & 97.67            \\
    +LP                   & 94.00       & 97.76      & 92.54       & 97.33       \\
    +LP+PA                & 93.70       & 97.89      & 92.17       & 97.42            \\
    +LP+PA+TG             & \textbf{94.68} & \textbf{98.42} & \textbf{92.62} & \textbf{97.69} \\
   \bottomrule
    \end{tabular}\label{fig:ab-modules}
\end{table}

\textbf{Sequential Training \textit{vs.} Joint Training} When using phenotypic saliency scores for prediction, PA-MIL$_J$ performs better on the in-domain test dataset, while PA-MIL$_S$ shows better out-of-domain performance. This suggests that separately training the phenotypic saliency prediction and the subtype classification can help the model achieve better out-of-domain generalization performance.

\subsection{Ablation Studies}
\textbf{Ablation of modules.} We conducted ablation experiments on the different modules of PA-MIL, and the experimental results are presented in Table~\ref{fig:ab-modules}. We replace the cross-attention mechanism that leverages language prompting with a self-attention mechanism to learn phenotypic features, creating a baseline.
Note that although baseline achieves good performance, it learns phenotype-agnostic features.
As shown in the table, language prompting, phenotypic feature alignment, and transcriptomic guidance each contribute to improving the model’s performance.
\begin{table}[thbp]
    \centering
    \caption{Ablation study of the guidance method, evaluated using mean accuracy and mean AUC across 5-fold on the NSCLC task. `Feat' represent feature-level guidance, and `Logit' means logit-level guidance.}
    \begin{tabular}{cccccc}
    \toprule
    \multirow{2}{*}{Feat} & \multirow{2}{*}{Logit} & \multicolumn{2}{c}{TCGA} & \multicolumn{2}{c}{CPTAC} \\ \cmidrule(lr){3-4} \cmidrule(lr){5-6} 
                       &                    & Acc         & AUC        & Acc         & AUC         \\ \midrule
                       &                    & 93.70       & 97.89      & 92.17  & 97.42            \\
    \checkmark         &                    & 94.67       & 97.90      & 92.53       & 97.59       \\
                       & \checkmark         & 94.49       & 97.90      & 92.30       & 97.24        \\
    \checkmark         & \checkmark         & \textbf{94.68} & \textbf{98.42} & \textbf{92.62} & \textbf{97.69} \\
   \bottomrule
    \end{tabular}\label{tab:ab-level}
\end{table}
\begin{table}[htbp]
    \centering
    \caption{Ablation study of the objective functions (Obj.), evaluated using mean accuracy and mean AUC across 5-fold on the NSCLC.}
    \begin{tabular}{ccccc}
    \toprule
    \multirow{2}{*}{Obj.} & \multicolumn{2}{c}{TCGA} & \multicolumn{2}{c}{CPTAC} \\ \cmidrule(lr){2-3} \cmidrule(lr){4-5} 
                          & Acc         & AUC        & Acc         & AUC         \\ \midrule
    CL                    & 93.52       & 97.84      & 91.38       & 97.31            \\
    L1                    & 94.68       & 97.98      & 92.14       & 97.64       \\
    L2                    & \textbf{94.68} & \textbf{98.42} & \textbf{92.62} & \textbf{97.69} \\
   \bottomrule
    \end{tabular}\label{tab:ab-obj}
\end{table}

\textbf{Effect of guidance methods.}
Table~\ref{tab:ab-level} presents the effect of different levels of guidance when training PA-MIL with transcriptomic supervision. The best performance is achieved by combining both feature-level and logit-level guidance, indicating that  multi-hierarchical supervision can effectively enhance PA-MIL’s learning.
\textbf{Effect of objective function.}
To facilitate PA-MIL learning knowledge from the transcriptomic data, we compared three objective functions: contrastive learning (CL)~\cite{tian2019contrastive}, L1 distance, and L2 distance.
Table~\ref{tab:ab-obj} illustrates that L2 distance exhibits the best performance.
\begin{figure}[thbp]
    \centering
    \begin{subfigure}[b]{0.23\textwidth}
        \centering
        \includegraphics[width=\textwidth]{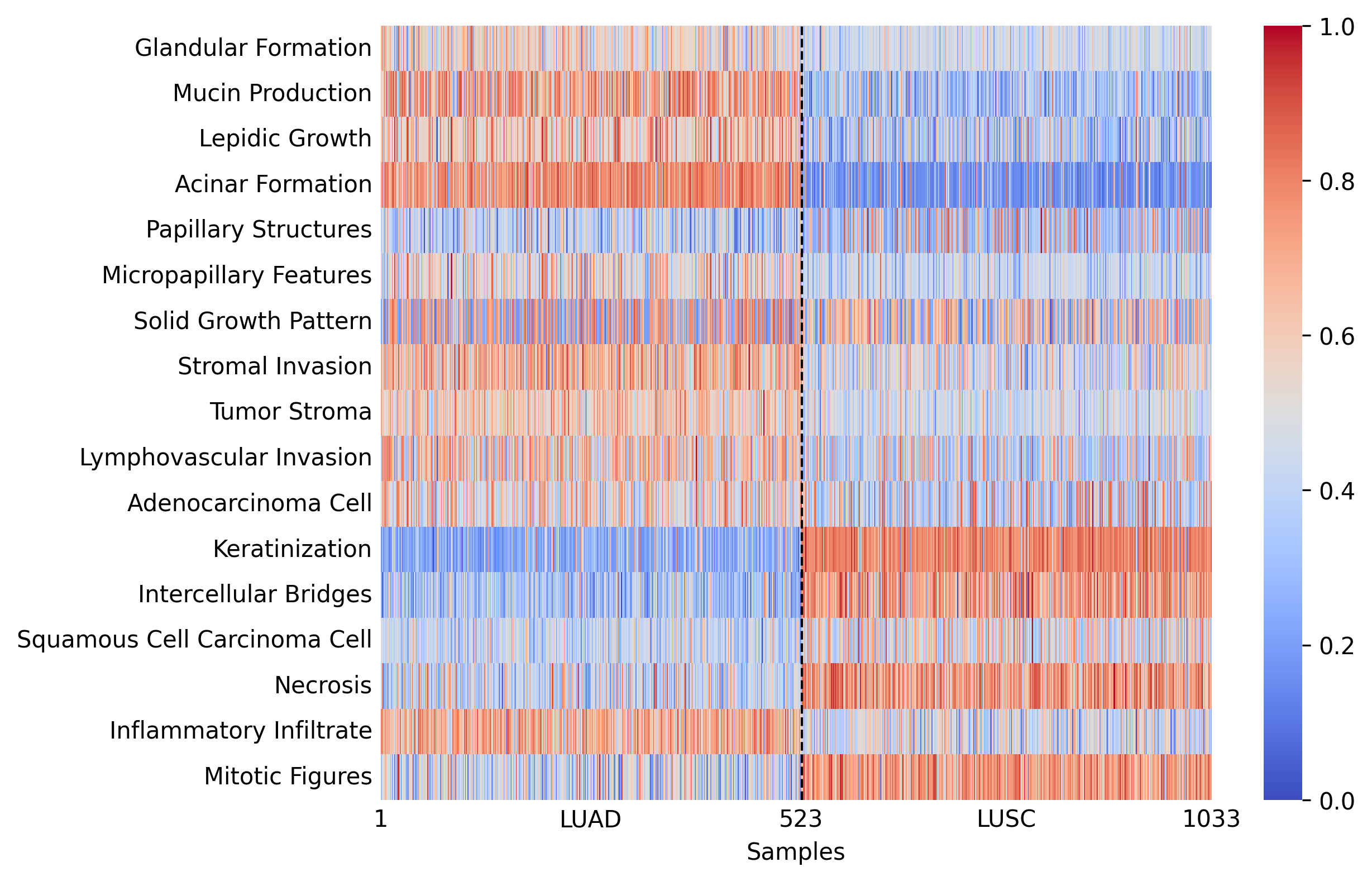}
    \end{subfigure}
    \hfill
    \begin{subfigure}[b]{0.23\textwidth}
        \centering
        \includegraphics[width=\textwidth]{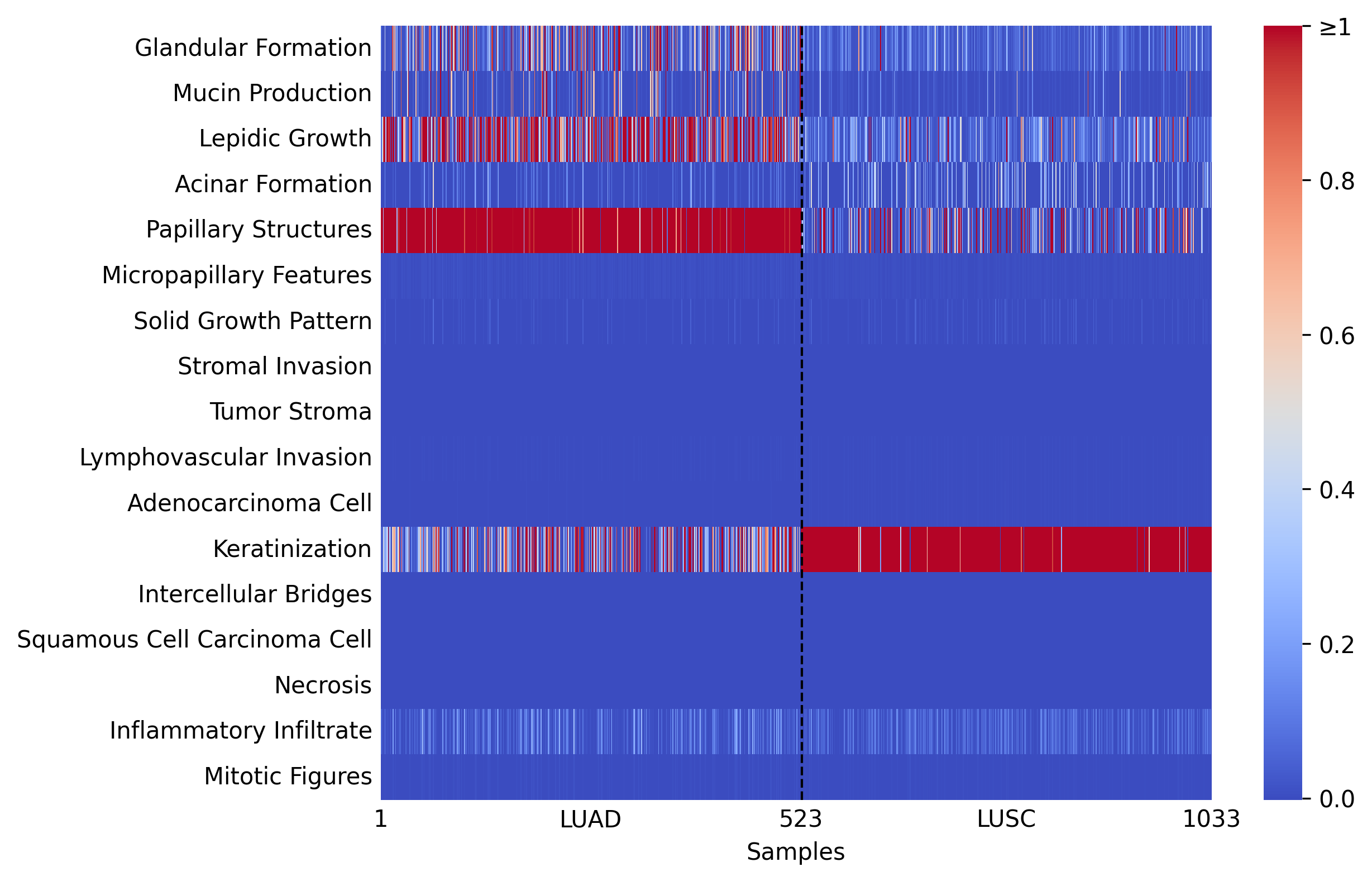}
    \end{subfigure}
    \caption{Heatmaps of phenotypic saliency scores under different activation functions. (\textbf{Left:} Layer normalization, \textbf{Right:} LeakyReLU) The heatmap illustrates the saliency scores of various phenotypes across 1372 samples from the CPTAC-NSCLC. Samples 1-683 are lung adenocarcinoma, and samples 684-1372 are lung squamous cell carcinoma. A higher value indicates a greater saliency score.}\label{fig:ablation_activation}
\end{figure}

\textbf{Ablation of activation functions for phenotype saliency scores.}
PA-MIL adopts an architecture similar to Concept Bottleneck Models (CBMs)~\cite{koh2020concept}, which are known to suffer from information leakage~\cite{margeloiu2021concept}.
To mitigate this issue, our method designs the phenotype saliency score not as a direct indicator of phenotype presence but as a normalized measure of relative phenotype importance within each sample. 
Specifically, layer normalization ensures that the mean and variance of saliency activations are standardized, redistributing any class-specific signals that might leak into a particular phenotype channel across all phenotypes. This activation function reduces the overall impact of the information leakage.
\Cref{fig:ablation_activation} illustrate the phenotype saliency scores on the test dataset under different activation functions. As shown, using LeakyReLU leads to significant information leakage, with class signals primarily leaking into “Papillary Structures” and “Keratinization.” In contrast, when layer normalization is applied, the saliency maps exhibit no observable leakage, confirming its effectiveness.

\section{Interpretability Analysis}
\subsection{Genotype-to-Phenotype Relationsips}
\label{sec:sec-in-gpnn}
In this section, we explore the interpretative analysis of the genotype-to-phenotype neural network to investigate the relationship between genotype and phenotype, as well as the relationship between phenotype and cancer subtype diagnosis.
Shapley values~\cite{shapley1953value} measure the marginal contribution of different input features to the prediction of a model.
\begin{figure}[thbp]
    \centering
    \includegraphics[width=0.47\textwidth]{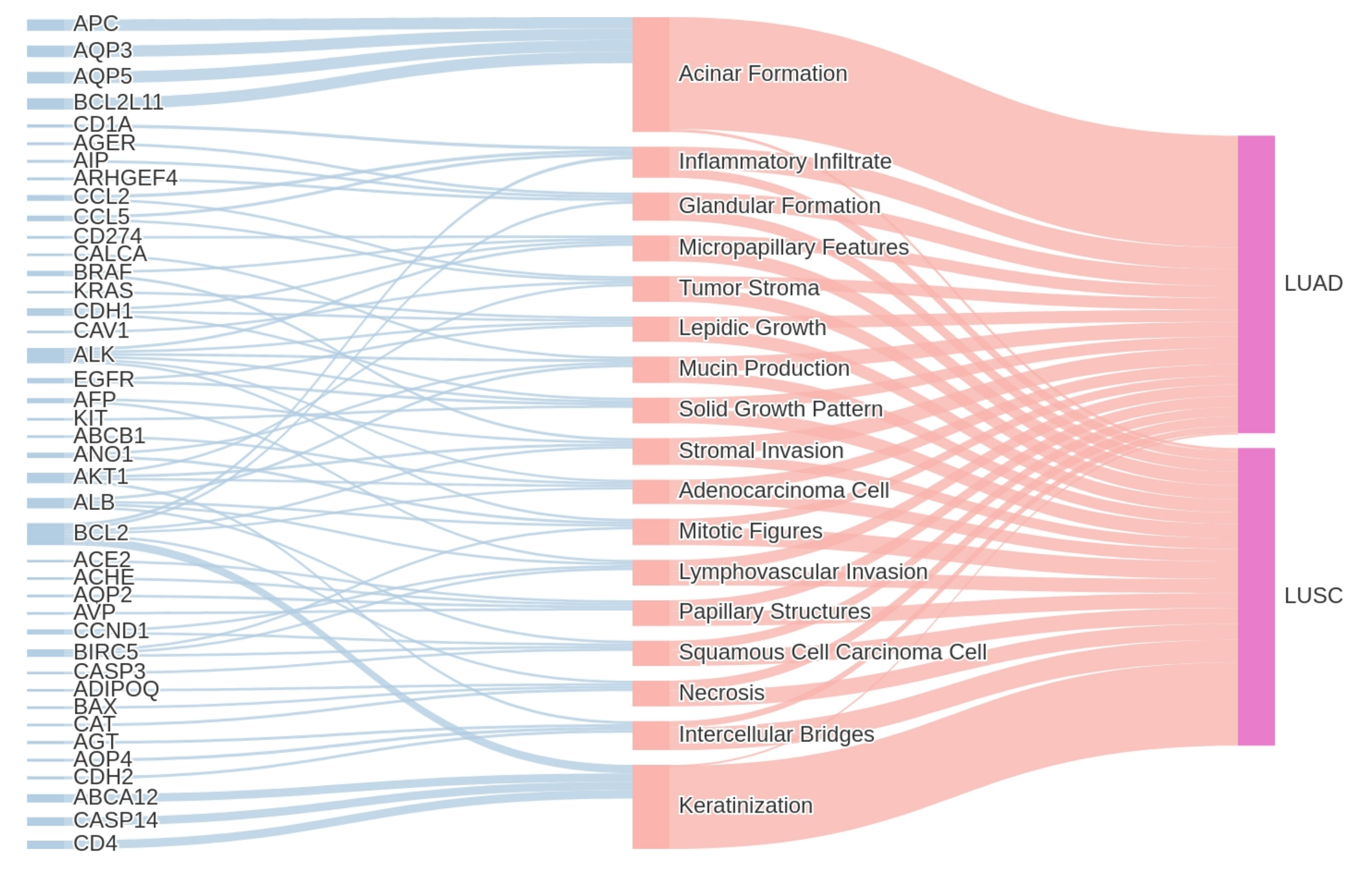}
    \caption{Relationships among genotypes, phenotypes, and cancer subtypes. Thicker lines means greater contributions. The top-4 genes contributing to each phenotype are shown.}
    \label{fig:int_sankey}
\end{figure}

\begin{figure}[thbp]
  \centering
  \begin{subfigure}[b]{0.22\textwidth}
    \centering
    \includegraphics[width=\textwidth]{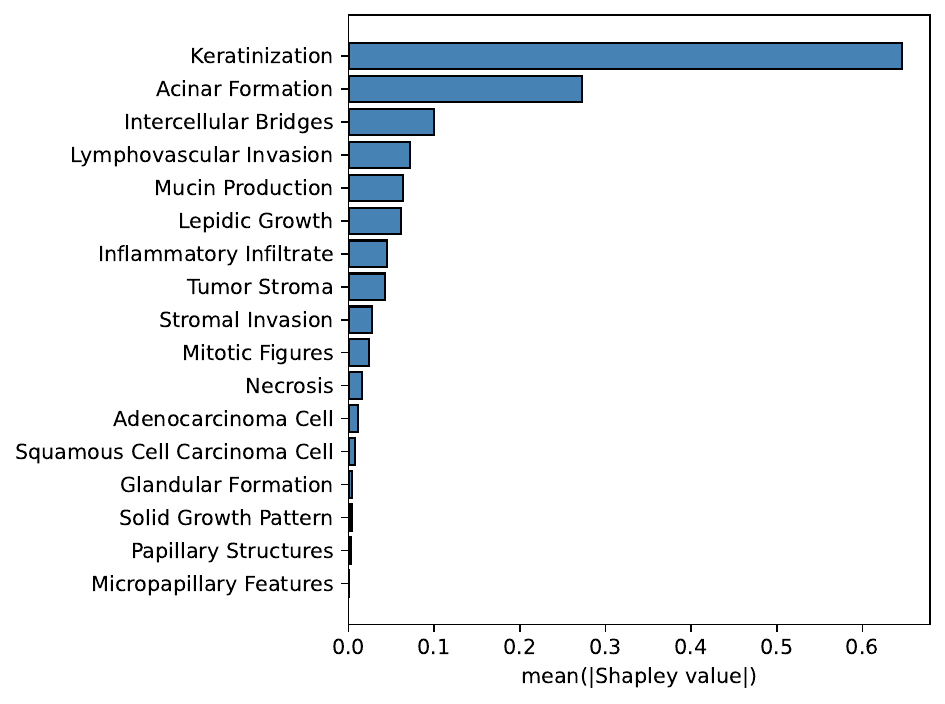}
    \subcaption{Ranking of phenotypic contributions to  subtyping.}\label{fig:int_global_a}
  \end{subfigure}
  \hfill
  \begin{subfigure}[b]{0.24\textwidth}
    \centering
    \includegraphics[width=\textwidth]{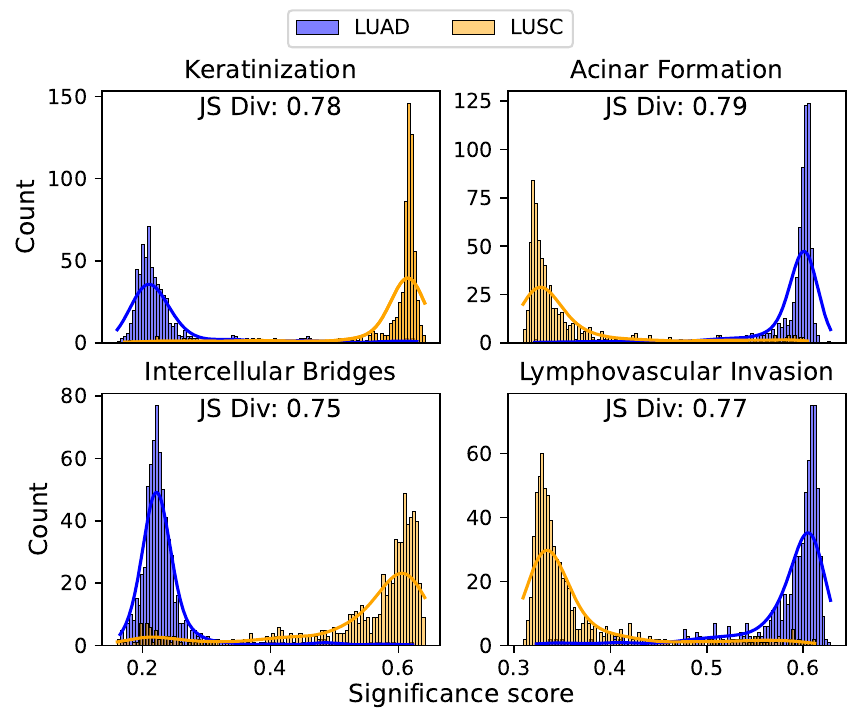}
    \subcaption{The saliency scores distribution of the top-4 contributing phenotypes.}\label{fig:int_global_b}
  \end{subfigure}
  \caption{Cohort interpretability analysis.}
\end{figure}
We use Shapley values to measure the contribution of gene expression levels to phenotypic saliency scores, as well as the contribution of phenotypic saliency scores to cancer subtype.
The visualization results are shown in Figure~\ref{fig:int_sankey}.
For example, the APC gene plays a crucial role in acinar formation. Studies~\cite{furlan2014apc} have shown that underexpression of APC leads to abnormal accumulation of $\beta$-catenin, which, via the Wnt signaling pathway, regulates cell proliferation and differentiation, ultimately contributing to the development of acinar cell carcinoma.
This demonstrates that by analyzing the GP-NN network, it is possible to explore the relationship between genotype and phenotype and generate new biological insights.
As shown in Figure~\ref{fig:int_sankey}, acinar formation and micropapillary structures contribute significantly to lung adenocarcinoma, while keratinization and intercellular bridges contribute significantly to lung squamous cell carcinoma.
These findings are consistent with the corresponding medical knowledge~\cite{nicholson20222021}: acinar formation, micropapillary structures, and keratinization are characteristic of the respective histological subtypes of lung adenocarcinoma and lung squamous cell carcinoma, while intercellular bridges are more prominent in squamous epithelium.



\begin{figure*}[thbp]
    \centering
    \begin{subfigure}[b]{0.96\textwidth}
        \centering
        \includegraphics[width=\textwidth]{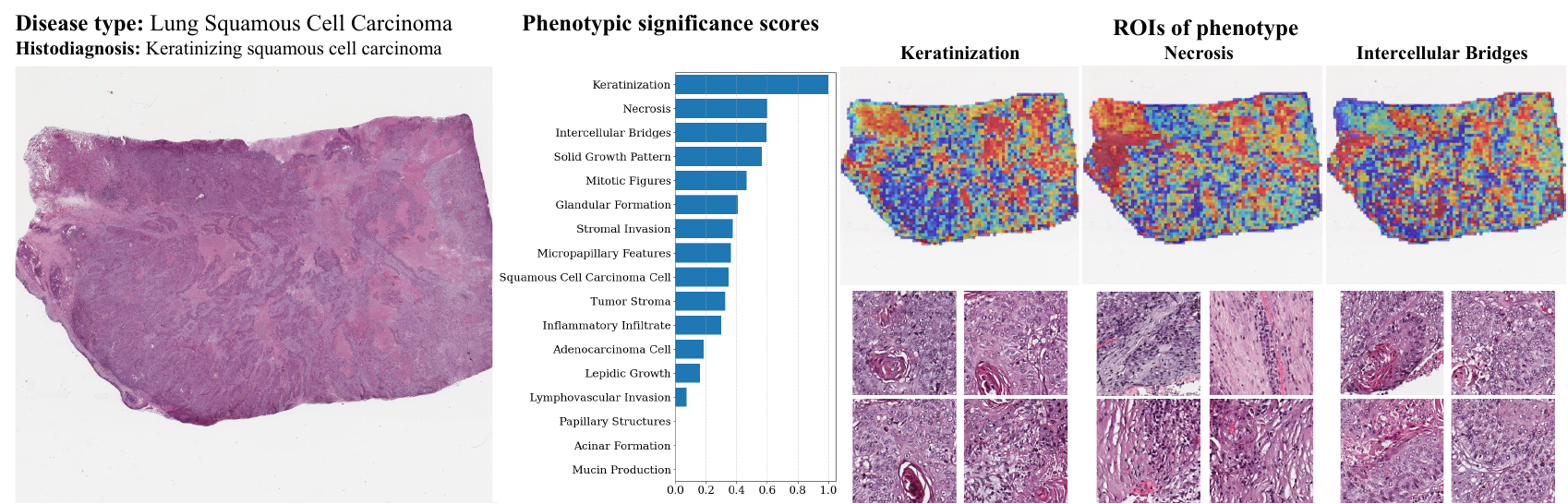}
    \end{subfigure}
    \vfill
    \begin{subfigure}[b]{0.96\textwidth}
        \centering
        \includegraphics[width=\textwidth]{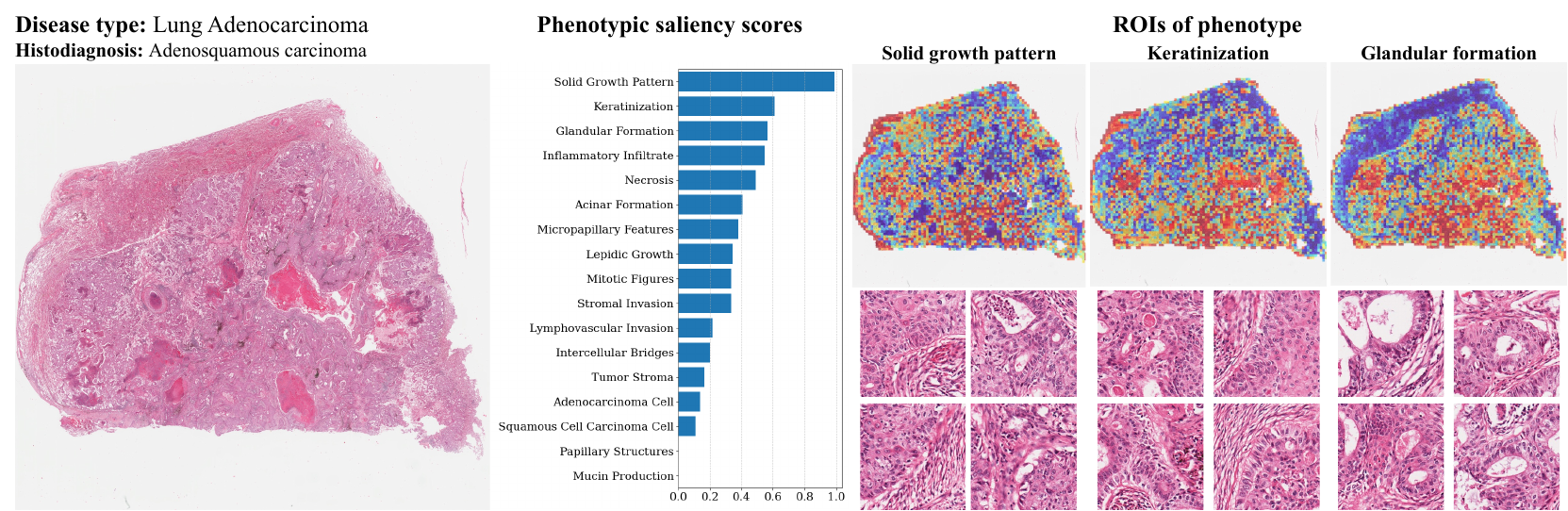}
    \end{subfigure}
    \caption{Case interpretability analysis. Every sample shows the prediction of phenotypic saliency scores, the heatmaps of the regions of interest for the top-3 contributing phenotypes, and representative patch images. \textbf{Top:} Keratinizing squamous cell carcinoma sample. \textbf{Bottom:}  Adenosquamous carcinoma sample. }\label{fig:int_case}
    \vspace{-2mm}
\end{figure*}
\subsection{Cohort Interpretability}
\label{sec:sec-in-global}
There are often common and specific patterns within a cohort and across cohorts.
We use the results of PA-MIL on the CPTAC-NSCLC to perform interpretability analysis from a cohort perspective.
Figure~\ref{fig:int_global_a} illustrates that keratinization and acinar formation are the most salient features for distinguishing lung adenocarcinoma from lung squamous cell carcinoma.
It can be observed that ``keratinization'' and ``acinar formation'' make the most significant contributions to the subtyping of lung adenocarcinoma and lung squamous cell carcinoma.
Figure~\ref{fig:int_global_b} shows the saliency distribution of the top-4 contributing phenotypes. We also use Jensen-Shannon (JS) divergence to measure the distance between the two distributions.
All four phenotypes exhibit high JS divergence values, indicating these phenotypes can effectively distinguish different subtypes.

\subsection{Case Interpretability}
\label{sec:sec-in-case}
We conduct a detailed analysis of the prediction results on the test samples from the perspective of individual case predictions.
Figure~\ref{fig:int_case} presents the prediction result for two samples.
It can be observed that for the sample histologically diagnosed as keratinized-type lung squamous cell carcinoma, PA-MIL accurately predicted keratinization as the most prominent phenotype and successfully localized the corresponding regions of interest.
In contrast, for the sample histologically diagnosed as acinar-type lung adenocarcinoma, PA-MIL can successfully predict that its most prominent phenotype is acinar formation and identify the corresponding regions of interest.
In practice, there are some adenosquamous carcinoma samples that contain both lung adenocarcinoma and lung squamous cell carcinoma. These samples are incorrectly labeled as lung adenocarcinoma or lung squamous cell carcinoma, making it challenging for single-label classification models trained on such data to recognize these complex patterns.
The bottom panel of Figure~\ref{fig:int_case} illustrates such a sample.
As shown in the figure, in the top-3 contributing phenotypes predicted by PA-MIL, features of squamous cell carcinoma, such as keratinization, and features of lung adenocarcinoma, such as glandular formation, are both present. This suggests that the phenotypes identified by PA-MIL can serve as evidence to recognize such complex patterns.

\section{Conclusion}
We present PA-MIL, a novel ante-hoc interpretable method for cancer subtyping.
PA-MIL identifies the saliency of cancer-related phenotypes from pathology images and subsequently performs cancer subtype diagnosis, providing reliable phenotypic evidence for diagnostic predictions.
Extensive experiments and interpretability analyses demonstrate the performance and reliability of PA-MIL.
Future work will explore learning more diverse phenotypes for application in various downstream tasks.

{
    \small
    \bibliographystyle{ieeenat_fullname}
    \bibliography{main}
}

\clearpage
\appendix
\renewcommand{\thefigure}{\Alph{section}.\arabic{figure}}
\renewcommand{\thetable}{\Alph{section}.\arabic{table}}
\setcounter{figure}{0}
\setcounter{table}{0}

\begin{center}
{\Large\bfseries APPENDIX}
\end{center}

\section{Dataset}
\textbf{NSCLC:} For the No-Small Cell Lung Carcinoma (NSCLC) subtyping task, we use the Hematoxylin and Eosin (H\&E) Formalin-Fixed and Paraffin-Embedded (FFPE) WSIs from TCGA and CPTAC. Both TCGA and CPTAC cohorts contain two subtypes: lung adenocarcinoma (LUAD) and lung squamous cell carcinoma (LUSC). The TCGA cohort contains a total of 1033 slides (LUAD: 523, LUSC: 510) and the CPTAC cohort contains a total of 1372 slides (LUAD: 683, LUSC: 689).

\textbf{RCC:} For the Renal Cell Carcinoma (RCC) subtyping, we use the H\&E-stained FFPE WSIs from TCGA and CPTAC. The TCGA cohort contains three subtypes: Clear Cell Renal Cell Carcinoma (CCRCC), Papillary Renal Cell Carcinoma (PRCC), and Chromophobe Renal Cell Carcinoma (CHRCC). The CPTAC cohort contains two subtypes: CCRCC and PRCC.
The TCGA cohort contains a total of 873 slides (CCRCC: 506, PRCC: 271, CHRCC: 66).
The CPTAC cohort contains a total of 555 slides (CCRCC: 524, PRCC: 31).

\section{Additional Experiment Results}
\textbf{Experiment results of GP-NN.} \Cref{tab:tab-gpnn} presents the experimental results of GP-NN for NSCLC subtyping and RCC subtyping based on transcriptomics data.
Transcriptomic data typically contain more distinctive biomarkers for differentiating cancer subtypes, and neural networks often achieve better performance in subtype classification when utilizing transcriptomic data.
From Table 1, it can be observed that PA-MIL, based on WSI data, achieves better performance in RCC subtyping.
 
\begin{table}[htbp]
    \caption{Experiment Results of GP-NN.}\label{tab:tab-gpnn}
    \resizebox{0.49\textwidth}{!}{
    \begin{tabular}{l|cc|cc}
    \hline
    \multirow{2}{*}{Method} & \multicolumn{2}{c|}{NSCLC} & \multicolumn{2}{c}{RCC} \\ \cline{2-5} 
                            & Acc.         & AUC         & Bal acc.      & F1      \\ \hline
    GP-NN(RNA-Seq)          & 97.19±1.53   & 98.82±0.64  & 95.13±1.07    & 95.59±1.26 \\
    PA-MIL(WSI)             & 94.68±1.30   & 98.42±0.84  & 95.66±1.86    & 95.69±0.59 \\ \hline
    \end{tabular}}
\end{table}

\begin{figure}[thbp]
    \centering
    \begin{subfigure}[b]{0.23\textwidth}
        \centering
        \includegraphics[width=\textwidth]{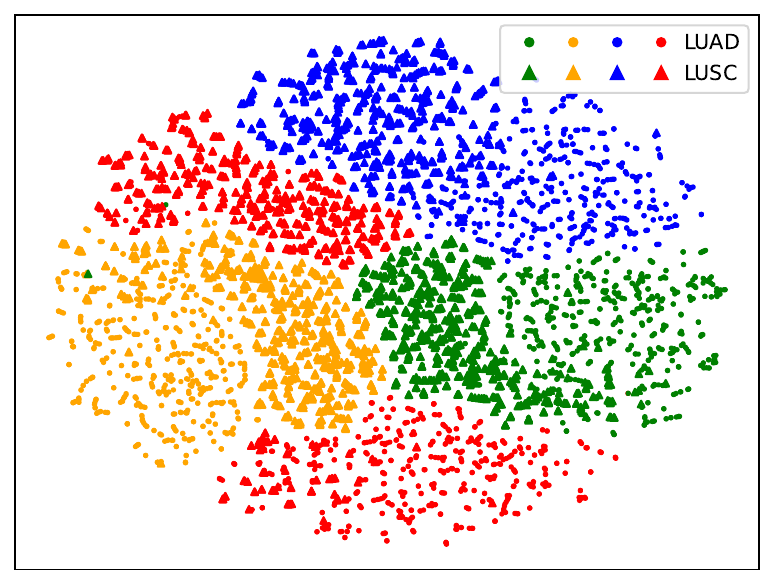}
        \caption{With PA}\label{fig:tsne1}
    \end{subfigure}
    \hfill
    \begin{subfigure}[b]{0.23\textwidth}
        \centering
        \includegraphics[width=\textwidth]{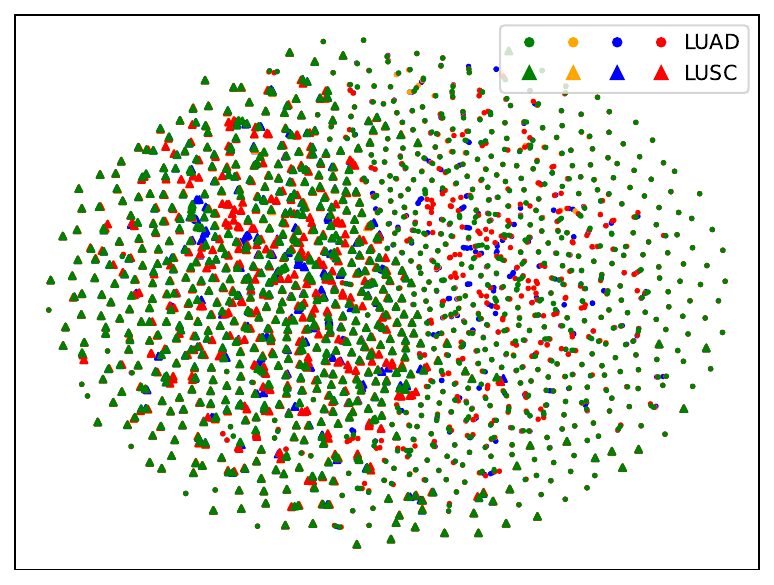}
        \caption{Without PA}\label{fig:tsne2}
    \end{subfigure}
    \caption{Distribution of phenotype features.}
    \label{fig:placeholder}
\end{figure}
\textbf{Ablation of phenotype features alignment (PA).}
 PA-MIL learns subtype-agnostic, mutually independent phenotypic features, enabling a concise and scalable solution for diverse cancer tasks. The primary goal of PA is to constrain phenotypic features, enhancing the reliability of model interpretation. 
 ~\Cref{fig:tsne1} and ~\Cref{fig:tsne2} show the distribution of features corresponding to the top-4 contributing phenotypes in the feature space, with PA enabled and disabled, respectively. As shown, in the absence of PA, the features are randomly scattered. In contrast, with PA enabled, the phenotypes form clearly distinct distributions.

\section{Phenotype Knowledge Base}
\Cref{tab:k-nsclc} and \Cref{tab:k-rcc} respectively present the phenotypic knowledge bases we have constructed for NSCLC and RCC.
The phenotypic knowledge base for NSCLC contains 17 phenotypes, while the phenotypic knowledge base for RCC contains 16 phenotypes.
Each phenotype includes its morphological description and potentially associated genes.
\onecolumn
\begin{longtable}{p{3cm}p{5.4cm}p{7cm}}
    \caption{Phenotype knowledge base of NSCLC.} \label{tab:k-nsclc} \\
        \toprule \textbf{Phenotype name} & \textbf{Morphological descirption} & \textbf{Associated gene sets} \\ \midrule
        \endfirsthead
        
        \multicolumn{3}{c}%
        {{\bfseries \tablename\ \thetable{} -- continued from previous page}} \\
        \toprule \textbf{Phenotype name} & \textbf{Morphological descirption} & \textbf{Associated gene sets} \\ \midrule
        \endhead
			
        \multicolumn{3}{l}{{Continued on next page}} \\ 
        \endfoot
			
        \bottomrule
        \endlastfoot
        
         Glandular Formation & The presence of gland-like structures lined by malignant cells that may produce mucin. These glandular patterns can appear as circular or irregular formations, often with luminal spaces containing mucin. & APC, MUC5AC, PTGS2, TP53, CTNNB1, KLK4, GHRH, PRL, ESR1, POMC, PTEN, FOXA2, EGFR, CD44, AKT1, MUC1, GAL, KRAS, MEN1, SOX9, PPARD, MAPK1, KCNJ5, VIM, AIP, CDH1, GHR, IGF1, MAP2K7, MYC, LGR5, PPARA, LIF, EZH2, IL6, CCND1, ARHGEF4, NFE2L2, FOXP3, KIT, AGER, CDK4, PAX8, BCL2, POU1F1, SPARC, FGF10, GJA1, CYP19A1, CLDN1 \\ \midrule
         Mucin Production & Cells within the tumor may contain intracytoplasmic mucin, which can be visualized as vacuoles inside the cells with special stains like mucicarmine or PAS with diastase. & MUC5AC, IL13, MUC2, EGFR, IL4, IL5, CFTR, MUC5B, ANO1, MUC1, TAC1, MUC3A, CXCL8, IL17A, VIP, MAPK1, CLCA1, GAST, IL10, TNF, IFNG, STAT6, SCT, TGFA, MPO, CALCA, ALB, TJP1, IL6, ALK, PTGS2, NOS2, OCLN, ELANE, HAVCR2, F2RL1, CDX2, IL1B, IGHE, LEP, CCL11, PTGER4, MUC4, CD79A, PTGER3, MMP9, CDH1, GCG, TLR4, IL22 \\ \midrule
         Lepidic Growth & Flattened or slightly atypical cells growing along pre-existing alveolar structures without invasion into the surrounding lung parenchyma. This pattern gives a paving stone appearance and is often seen in the early stages of adenocarcinoma in situ. & EGFR, SERPINB3, MUC5AC, ALK, PDPN, TP63, CD274, VEGFA, CDH1, KRAS, NKX2-1 \\ \midrule
         Acinar Formation & This pattern consists of round or ovoid acini or tubules made up of neoplastic cells, resembling normal lung acinar structures. & KRAS, EGFR, CCK, AQP5, MUC5AC, ALK, TAC1, BHLHA15, CDH1, CD274, TNF, SOX9, TP53, MMP9, GCG, BCL2L11, FCGRT, TP63, VIP, NR5A2, SST, GRP, SNAI1, MAP1LC3A, PDX1, MUC1, PIK3CA, TLR3, NEUROG3, SMARCA4, VMP1, RHOA, VAMP2, APC, MAPK1, KIT, KRT19, CDK2, GRB2, CD34, MYC, KRT7, SCT, CXCL2, CTNNB1, BRAF, AQP3, STAT3, GPX4, TIMP1 \\ \midrule
         Papillary Structures & The presence of finger-like projections lined by neoplastic cells with or without fibrovascular cores. These projections can be complex and branching. & MUC5AC, CALCA, BDNF, GNAT3, TAC1, CD36, SHH, TAS1R3, UCHL1, TCHH, SOX2, TAS1R2, TAS2R38, TAS1R1, BMP2, ENO2, TP53, IGF1, AVP, TRPV1, FFAR4, BRAF, EGFR, KRT7, ACHE, NTRK2, VIM, TP63, BMP4, TGFB1, CD34, ACE2, KRT14, NGFR, AQP2, SNAP25, MME, GAL, CD79A, NPY, LGR5, SPP1, NTF3, NCAM1, GLI1, SST, FGF2, FST, FLG, MKI67 \\ \midrule
         Micropapillary Features & Small papillary tufts lacking a fibrovascular core, often appearing as free-floating within alveolar spaces, and associated with high-grade tumors and poor prognosis. & EGFR, ERBB2, TP53, MUC1, MUC5AC, ALK, CDH1, NKX2-1, CD274, TP63, BRAF, WT1, KRT7, MME, PAX8, SMARCA2, KRAS, MET \\ \midrule
         Solid Growth Pattern & Sheets or nests of cells with little or no glandular or acinar differentiation. There can be high cellularity, and necrosis may be present. & EGFR, ALK, TP53, MUC5AC, RET, CDH1, KRAS, ACCS, PTGS2, PCNA, MKI67, TG, VEGFA, KIT \\ \midrule
         Stromal Invasion & Invasive adenocarcinoma is characterized by the neoplastic cells breaching the basement membrane and infiltrating into the surrounding stroma, with possible desmoplastic reaction. & MMP9, AKT1, CDH1, MMP2, ERBB2, VEGFA, EGFR, TP53, MAPK1, STAT3, HGF, CXCR4, CD44, SNAI1, VIM, PTGS2, MMP14, MET, PTK2, PGR, PIK3CA, TGFB1, IL6, PTEN, AFP, MTOR, CXCL12, MAPK3, ZEB1, CDH2, PLAUR, SRC, PLAU, CD274, MYC, TWIST1, BCL2, HPSE, RAC1, EZH2, RHOA, BSG, CCND1, MMP7, MMP1, CXCL8, YAP1, NFKB1, CTNNB1, BRAF \\ \midrule
         Tumor Stroma & A variable amount of connective tissue can accompany the tumor cells, ranging from a scant fibrous stroma to dense fibrosis, which is commonly seen in the scarred or fibrotic focus of the tumor. & CD274, PDCD1, VEGFA, CD8A, IL6, EGFR, STAT3, FOXP3, TP53, IDO1, IL10, CXCR4, CD4, CCL2, ERBB2, CTLA4, TGFB1, FAP, CXCL12, METTL3, CD68, PTGS2, AKT1, MMP9, MMP2, CD163, CXCL8, TNF, IL17A, IL2, KRAS, NT5E, CD44, HIF1A, MTOR, HGF, CAV1, IFNG, MYC, HAVCR2, CCL5, CD47, VIM, POSTN, PIK3CA, PTEN, SPARC, CSF2, MUC5AC, SPP1 \\ \midrule
         Lymphovascular Invasion & Tumor cells may be seen within lymphatic or blood vessels, suggesting a mechanism for potential metastasis. & VEGFA, ERBB2, TP53, AFP, PGR, MMP9, CDH1, EGFR, CD274, MMP2, ALB, BCL2, PTGS2, BRAF, MUC5AC, PDPN, PECAM1, CD34, CEACAM5, CCND1, SERPINB3, KRAS, KDR, CRP, CD44, AKT1, VEGFC, NODAL, BIRC5, CDKN2A, MKI67, COX8A, PCNA, HIF1A, MUC1, FGF2, PROM1, VIM, CXCR4, MET, CTNNB1, PTEN, PIK3CA, MUC16, FLT1, IL6, NFKB1, LGALS3, VEGFD, SPP1 \\ \midrule
         Adenocarcinoma Cell & The lung adenocarcinoma cells include pleomorphism of the nucleus with varying sizes, high nuclear to cytoplasmic ratio, active nuclear division, abundant cytoplasm, commonly exhibiting glandular structures, tight cell arrangement, and minimal intercellular stroma. & SERPINB3, EGFR, TP53, CD274, MUC5AC, KRAS, ALK, ERBB2, AKT1, CDH1, MALAT1, TP63, PTGS2, CDKN2A, PDCD1, NKX2-1, VEGFA, MAPK1, BCL2, MUC1, CD44, CEACAM5, TGFB1, ABCB1, EPCAM, PCNA, MYC, CCND1, IFNG, CDKN1A, TNF, STAT3, CASP3, IL6, PIK3CA, NFKB1, TNFSF10, PROM1, MTOR, ENO2, KRT5, MET, MMP9, BIRC5, BRAF, BAX, MMP2, VIM, CD8A, PTEN \\ \midrule
         Keratinization & Keratinization is indicated by the presence of orangeophilic (eosinophilic), dense, lamellar keratin within cancer cells or as keratin pearls. Keratin pearls are formed when concentric layers of squamous cells undergo keratinization resulting in a whorled appearance. & FLG, IVL, KRT10, SERPINB3, TGM1, KRT1, CASP14, TP53, TCHH, KRT14, MUC5AC, KRT16, SPRR1B, KRT13, HRNR, CDKN2A, KRT19, EGFR, STS, KRT17, TGM3, BCL2, GJB2, CDSN, NFE2L2, OMA1, CST6, KRT5, GJB4, KLK7, FGF7, ABCA12, CD8A, MMP9, VIM, KRT2, PRL, LPA, NUTM1, GJA1, KLRD1, HSPB1, CD4, CLU, IL17A, ATP2A2, IL2, KLRB1, GJB3, CTNNB1 \\ \midrule
         Intercellular Bridges & Also known as desmosomes, intercellular bridges appear as small lines or tiny spicules joining adjacent tumor cells. They are best viewed at high magnification and are indicative of squamous differentiation. & GJA1, GJD2, GJB1, GJB2, GJA5, GJC1, GJB6, GJA8, GJA4, PANX1, TJP1, GJA3, CDH2, CDH1, GJC2, SRC, OCLN, MAPK1, EMILIN1, MIP, REN, TEX14, PVALB, DSP, AKT1, GJB3, VEGFA, OXT, EDN1, TNF, AGT, CDH5, AQP4, KNG1, IGF1, MAPK8, CTNNB1, CGA, IL1B, VIM, PRKCG, SCN5A, KCNA3, KIT, GFAP, GJB4, GNRH1, NEUROD1, DES, MAPK3 \\ \midrule
         Squamous Cell Carcinoma Cell & Squamous cell carcinoma cells are often polygonal in shape and show varying degrees of atypia, including large hyperchromatic nuclei, prominent nucleoli, and irregular nuclear contours. The cytoplasm can range from scant to eosinophilic and dense, sometimes with clear perinuclear halos. & SERPINB3, EGFR, TP53, CDKN2A, CD274, AKT1, VEGFA, PDCD1, CCND1, CDH1, TP63, PTGS2, STAT3, BCL2, PIK3CA, ERBB2, CD44, BIRC5, CEACAM5, MMP9, IL6, KRT19, MTOR, MAPK1, MYC, CDKN1A, ENO2, PTEN, SOX2, MMP2, PDPN, NODAL, ALB, PCNA, MET, DNAJB7, CXCR4, VIM, YAP1, IL2, CASP3, CXCL8, MUC1, CD8A, CRP, KRAS, CTNNB1, NOTCH1, NFE2L2, TGFB1 \\ \midrule
         Necrosis & Necrosis can often be seen as areas of more eosinophilic staining where the tissue architecture is lost. Geographic necrosis, where large areas necrose in a map-like pattern, may be present. & TNF, IL6, IL1B, IL10, IFNG, CRP, CXCL8, CCL2, IL2, IL4, TNFSF10, NFKB1, PTGS2, IL17A, NOS2, VEGFA, ICAM1, CASP3, CSF2, IL1A, MPO, ADIPOQ, LEP, TNFRSF1A, FAS, TLR4, TGFB1, SOD1, BCL2, IL18, TNFRSF1B, GPT, VCAM1, MMP9, IL5, CAT, IL1RN, IL13, RELA, TP53, LTA, MAPK1, MAPK14, BAX, HMGB1, FASLG, MAPK8, IFNA1, AKT1, LPA \\ \midrule
         Inflammatory Infiltrate & A variable amount of inflammatory cells, including lymphocytes and macrophages, can commonly be found infiltrating the tumor stroma, sometimes obscuring the tumor cells. & TNF, CD68, CD4, ICAM1, IFNG, CD8A, CCL2, IL10, IL17A, MPO, IL4, VEGFA, FOXP3, MUC5AC, IL1B, PTGS2, IL6, IL5, TGFB1, IL2, NOS2, MMP9, VCAM1, PTPRC, BCL2, NFKB1, CCL5, IL13, CASP3, CSF2, CD1A, TNFRSF8, MS4A1, CD34, ITGAM, PRF1, CD274, FASLG, TP53, GZMB, PDCD1, TLR4, CXCL2, IGHE, PECAM1, CXCL1, MPV17, ITGAX, KIT, ELN \\ \midrule
         Mitotic Figures & Mitotic figures indicate cellular division. These figures can appear as dark, dense, thread-like structures within the nucleus of cells undergoing mitosis. & TP53, MKI67, PCNA, MUC5AC, KIT, CD34, BCL2, CDKN2A, VIM, GFAP, DES, EGFR, BRAF, SYP, BRCA1, CCND1, CDKN1A, VEGFA, IDH1, BIRC5, ENO2, MAD2L1, CDK1, ALK, MYC, CD99, KRT7, CDH1, KRT19, ERBB2, EZH2, TP63, PECAM1, SERPINB3, TERT, CSF3, PRL, ALB, CASP3, SST, IGF1, CHGA, EPO, CCL4, IL2, TOP2A, ANO1, MLANA, BAP1, CTNNB1
\end{longtable}
\begin{longtable}{p{3cm}p{5.4cm}p{7cm}}
    \caption{Phenotype knowledge base of RCC.} \label{tab:k-rcc} \\
        \toprule \textbf{Phenotype name} & \textbf{Morphological descirption} & \textbf{Associated gene sets} \\ \midrule
        \endfirsthead
        
        \multicolumn{3}{c}%
        {{\bfseries \tablename\ \thetable{} -- continued from previous page}} \\
        \toprule \textbf{Phenotype name} & \textbf{Morphological descirption} & \textbf{Associated gene sets} \\ \midrule
        \endhead
			
        \multicolumn{3}{l}{{Continued on next page}} \\ 
        \endfoot
			
        \bottomrule
        \endlastfoot

        Clear Cytoplasm & Tumor cells have a clear or vacuolated cytoplasm due to the accumulation of lipids or glycogen, presenting as pale or unstained cytoplasm. & MUC5AC, MME, VIM, AFP, KRT7, TFE3, MLANA, KIT, TP53, CD68, EWSR1, SALL4, BCL2, GPC3, DES, GFAP, PLIN2, CALCA, CA9, KRT20, HSPA4, NCAM1, MS4A1, MUC1, PRCC, TP63, POU5F1, SDHB, TFEB, NR5A1, TAC1, KRT8, TGFB1, CDX2, TJP2, SPOP, SERPINB3, VHL, ELOC, TJP1, ENO2, CDKN2A, HMGB1, CTNNB1, SYP \\ \midrule
        Eosinophilic Cytoplasm & The cytoplasm is more granular and exhibits eosinophilia, typically appearing pale pink or brown due to the presence of a higher number of mitochondria and other organelles within the cell. & MUC5AC, VIM, CD68, DES, KIT, TFE3, SYP, PRCC, ENO2, KRT7, AFP, KRT20, CD163, SOX10, MB, S100B, CD34, CALB2, MLANA, CDKN2A, CHGA, TP63, ALK, BAP1, SDHB, SMARCB1, CD99, EWSR1, BRAF, CASP3, SPN, MPO, GPC3, EPX, WT1, MUC1, GFAP, TNFRSF8, CLC, GATA3, RNASE3, MS4A1, AMACR, PIP, TP53, SETD2, PECAM1, EGFR, CDH1, SST \\ \midrule
        Papillary Structure & The cancer tissue shows typical papillary structures, with a central vascular core, and cells arranged in clusters or layers, with broad bases. & MUC5AC, RET, PTCH1, BRAF, TPCN1, VEGFA, TG, CDH1, CD34, KRAS, FGFR3, NTRK1, PECAM1, SPP1, ESR1, KNG1, TP63, CCDC6, EFEMP1, EGFR, SHCBP1 \\ \midrule
        Capsular Invasion & Tumor cells may invade through the renal capsule, extending into the perirenal fat tissue, leading to an ill-defined tumor margin. & BRAF, AFP, TP53, TG, VEGFA, ALB, CALCA, KRT19, COX8A, CDH1, RET, PTEN, CD274, ERBB2, CTNNB1, TERT, MMP2, CCND1, LGALS3, KRT7, BCL2, EGFR, PTGS2, TPO, CASP3, SLC2A5, CCDC6, TSHR, TIMP2, HIF1A, MUC5AC, PPARG, PCNA, FOLH1, PTH, GPT, HSPA4, CD44, MKI67, NOTCH1, EBAG9, CRP, MDM2, NODAL, MME, ALPP, CEACAM5, CD34, VHL, AGGF1 \\ \midrule
        Vascular Proliferation & The tumor exhibits numerous irregular blood vessels with thin walls, often showing areas of hemorrhage or thrombus, creating a vascular network appearance. & VEGFA, KDR, FGF2, AKT1, PECAM1, HIF1A, PTGS2, MAPK1, MMP9, EGFR, FLT1, TP53, ANGPT2, COL18A1, STAT3, THBS1, CD34, ANGPT1, MTOR, IL6, CXCL8, TEK, TGFB1, HGF, MMP2, CXCL12, PIK3CA, ENG, CXCR4, PGF, NOS3, AGT, BCL2, MAPK3, CCL2, TNF, IGF1, SERPINF1, ANG, NRP1, NFKB1, HPSE, EPO, LEP, CDH5, ERBB2, SPP1, CASP3, MYC, DLL4 \\ \midrule
        Large and Irregular Nuclei & Cancer cells have large, irregularly shaped nuclei, often with prominent nucleoli. & THPO, IL3, CD34, MPL, CSF2, KITLG, IL6, GATA1, ITGA2B, EPO, PF4, CSF3, IL11, ITGB3, MSLN, PTPN4, NFE2, FLT3, JAK2, RUNX1, ACHE, TGFB1, GP1BA, FLI1, AKT1, GATA2, KIT, MAPK1, ZFPM1, GP6, TIMP1, PPBP, SPI1, CXCL12, PTPN11, TAL1, CALR, MYB, TP53, CXCR4, IL1B, BCL2L1, GFI1B, FLT3LG, SPP1, PRG4, ALB, IFNA1, KLF1, CASP3 \\ \midrule
        Capillary Network & A rich capillary network is often observed within the tumor, particularly surrounding the papillary structures. & VEGFA, EDN1, ALB, AGT, TNF, NOS3, ICAM1, KNG1, KANTR, PECAM1, REN, NOS2, CALCA, LPA, CD34, ACE, KDR, ADAMTS13, IL6, EGR3, SERPINC1, PLAT, HMOX1, PROC, VCAM1, SOD1, PTGS2, AKT1, AVP, IL2, GAST, SST, FGF2, VIP, ANGPT1, GCG, MPO, SDC1, EPO, THBD, NPY, KLK4, ADIPOQ, ITGB2, TAC1, IGF1, NOS1, HIF1A, TGFB1, EDNRA \\ \midrule
        Nuclear-to-Cytoplasm Ratio & The nuclei occupy a smaller proportion of the cell volume, with relatively abundant cytoplasm. & CRP, ALB, LPA, MAPK8, CST3, IL6, BAX, CASP3, BCL2, MAPK1, JUN, NAGLU, APOA1, GDF15, PPP1R1A, APOB, PROC, TNF, AKT1, MAPK3, EAF2, ABCG2, BGLAP, TP63, MET, CYP2C19, NPPB, LCN2, PROS1, MAPK14, RBP4, CASP9, REN, LEP, ADIPOQ, CAT, MPO, EGFR, MTOR, RELA, TF, TP53, IGF1, GPT, SOD1, TLR4, CYP2D6, SERPINC1, IL10, PARP1 \\ \midrule
        Glandular Structure & Some tumors exhibit glandular patterns, with tubular or acinar arrangements of cells. & MUC5AC, KLK4, KLK2, CDH1, REN, CEACAM5, LIF, GAST, PTGS2, VIM, CD44, ADIPOQ, PTH \\ \midrule
        Inflammatory Infiltrate & A variable amount of inflammatory cells, including lymphocytes and macrophages, can commonly be found infiltrating the tumor stroma, sometimes obscuring the tumor cells. & CCL2, TNF, IL10, MPO, IL17A, IL4, CD68, IL6, ICAM1, CD4, IL13, PTGS2, VEGFA, IFNG, CD8A, TLR4, IL5, NLRP3, IL1B, MMP9, NOS2, MUC5AC, PTPRC, VCAM1, FOXP3, IL2, BCL2, SOD1, NFKB1, TGFB1, CXCL2, CCL5, ALB, RELA, IL18, CASP3, HMGB1, CD274, PECAM1, CSF2, GPT, STAT3, MAPK1, MYD88, PDCD1, IL23A, CASP1, CAT, AKT1, CD34 \\ \midrule
        Vascular Endothelial Proliferation & The presence of numerous small, thin-walled blood vessels interspersed among tumor cells, forming a network-like pattern. & VEGFA, KDR, FGF2, AKT1, PECAM1, HIF1A, PTGS2, MMP9, EGFR, FLT1, ANGPT2, COL18A1, MAPK1, TP53, THBS1, STAT3, CD34, ANGPT1, MTOR, IL6, CXCL8, TEK, HGF, TGFB1, MMP2, CXCL12, ENG, PIK3CA, CXCR4, NOS3, PGF, BCL2, CCL2, TNF, MAPK3, SERPINF1, IGF1, ANG, NRP1, HPSE, EPO, NFKB1, LEP, CDH5, DLL4, CASP3, ERBB2, MET, SPP1, MYC \\ \midrule
        Granular Cell Type & Cells containing abundant eosinophilic granules in the cytoplasm, resulting in a granular appearance. & FCGR3A, NCAM1, ENO2, B3GAT1, VIM, PRF1, CD8A, CALB2, TIA1, MS4A1, CD2, TGFBI, FOS, STAT3, GFAP, CD68, CCK, IL2, MUC1 \\ \midrule
        Necrosis & Areas within the tumor exhibiting necrotic tissue, appearing as irregular, acellular regions. & TNF, IL6, IL1B, IL10, IFNG, PDCD1, CASP3, CD274, CRP, BCL2, TP53, IL2, CXCL8, CCL2, IL4, FAS, TNFSF10, NFKB1, BAX, PTGS2, VEGFA, IL17A, NOS2, AKT1, MAPK1, ICAM1, CSF2, MAPK8, SOD1, MPO, FASLG, ADIPOQ, IL1A, TNFRSF1A, CASP8, TLR4, LEP, TGFB1, IL18, PARP1, MAPK14, CAT, ANXA5, CASP1, CTLA4, TNFRSF1B, HMGB1, GPT, VCAM1, MMP9 \\ \midrule
        Fibrous Stroma & Presence of dense fibrous tissue within the tumor, often surrounding blood vessels or tumor nests. & CD34, MUC5AC, KRT19, TGFB1, ELN, KRT7, CD44, CD68, VEGFA, VIM, MME, EPCAM, S100A4, SYP, CDH1, DES, CD274, GFAP, DCN, FGF2, TNFSF11, MUC1, FAP, NES, MFAP5, STAB1, PDGFRB, PGR, SALL4, CDKN2A, THBS1, TP53, LOXL2, PECAM1, PDCD1, MMP2, CD4, MKI67, FGF7, CALCA, DPP4, TIMP1, F13A1, UCHL1 \\ \midrule
        Mitotic Activity & The presence of numerous mitotic figures, indicating active cell division, with visible separation of nuclear and cytoplasmic components. & TP53, CDK1, PCNA, MKI67, KIT, PLK1, BCL2, ERBB2, CCNB1, PGR, CD34, VEGFA, WEE1, CDKN2A, VIM, CDC25C, PTPA, CDKN1A, KIF11, CCND1, CDC20, EGFR, AURKA, CDC14A, CASP3, MYC, CAT, MUC5AC, CDH1, TTK, BUB1B, MAPK1, MAD2L1, BIRC5, PTGS2, PRL, FGF2, IGF1, TOP2A, FZR1, CDK2, AKT1, APC, AURKB, IL6, BRCA1, MTOR, BRAF, ALB, KIF2C \\ \midrule
        Calcification & Focal calcification within the tumor, which appears as basophilic or eosinophilic deposits, often seen as small granular or block-like structures. & BGLAP, RUNX2, SPP1, ALPP, BMP2, PTH, MGP, TNFRSF11B, FGF23, IBSP, AHSG, SP7, LPA, ALPL, DMP1, DSPP, ENPP1, ELN, TNFSF11, KL, MAPK1, CRP, ABCC6, SOST, SPARC, AKT1, ALB, VDR, PHEX, SLC20A2, VEGFA, IGF1, TGFB1, IL6, COL1A1, FGF2, CCL27, BMP7, BMP4, LEP, AMELX, MAPK3, SOX9, GDF2, CALCA, MSX2, ANKH, MAPK14, APOE, TNF \\
\end{longtable}
\twocolumn

\end{document}